\documentclass[letterpaper, 10 pt, conference]{ieeeconf}  

\IEEEoverridecommandlockouts                              

\usepackage{hyperref}
\hypersetup{
  colorlinks=true,
  urlcolor=blue,   
  linkcolor=blue,  
  citecolor=blue   
}

\usepackage{amsmath,amssymb,amsfonts}
\usepackage{dsfont}
\newcommand{\ind}{\mathds{1}}  

\usepackage{graphicx}

\usepackage{booktabs}
\usepackage{multirow}

\usepackage[table]{xcolor}
\definecolor{mygreen}{rgb}{0.7, 0.9, 0.7}
\definecolor{myyellow}{rgb}{1.0, 1.0, 0.7}

\usepackage{soul}
\sethlcolor{myyellow}

\usepackage{cite}

\newcommand{\best}[1]{\cellcolor{mygreen}\textbf{#1}}
\newcommand{\second}[1]{\cellcolor{myyellow}#1}

\usepackage[font=small,labelfont=bf,labelsep=colon]{caption}

\title{\LARGE \bf
OpenMonoGS-SLAM: Monocular Gaussian Splatting SLAM\\
with Open-set Semantics
}

\author{
    Jisang Yoo$^{1}$,
    Gyeongjin Kang$^{1}$,
    Hyun-kyu Ko$^{1}$,
    Hyeonwoo Yu$^{1}$,
    Eunbyung Park$^{2}$%
    \thanks{$^{1}$Sungkyunkwan University, Suwon, Republic of Korea.}%
    \thanks{$^{2}$Yonsei University, Seoul, Republic of Korea. Corresponding author: Eunbyung Park.}%
}

\IEEEaftertitletext{%
\vspace{-0.6\baselineskip} 
\begin{center}
\footnotesize \url{https://jisang1528.github.io/OpenMonoGS-SLAM/}
\end{center}
}

\begin{document}

\maketitle
\thispagestyle{empty}
\pagestyle{empty}

\begin{abstract}

Simultaneous Localization and Mapping (SLAM) is a foundational component in robotics, AR/VR, and autonomous systems.
With the rising focus on spatial AI in recent years, combining SLAM with semantic understanding has become increasingly important for enabling intelligent perception and interaction.
Recent efforts have explored this integration, but they often rely on depth sensors or closed-set semantic models, limiting their scalability and adaptability in open-world environments.
In this work, we present OpenMonoGS-SLAM, the first monocular SLAM framework that unifies 3D Gaussian Splatting (3DGS) with open-set semantic understanding.
To achieve our goal, we leverage recent advances in Visual Foundation Models (VFMs), including MASt3R for visual geometry and SAM and CLIP for open-vocabulary semantics.
These models provide robust generalization across diverse tasks, enabling accurate monocular camera tracking and mapping, as well as a rich understanding of semantics in open-world environments.
Our method operates without any depth input or 3D semantic ground truth, relying solely on self-supervised learning objectives.
Furthermore, we propose a memory mechanism specifically designed to manage high-dimensional semantic features, which effectively constructs Gaussian semantic feature maps, leading to strong overall performance.
Experimental results demonstrate that our approach achieves performance comparable to or surpassing existing baselines in both closed-set and open-set segmentation tasks, all without relying on supplementary sensors such as depth maps or semantic annotations.

\end{abstract}

\section{INTRODUCTION}
\label{sec:intro}

Simultaneous Localization and Mapping (SLAM) plays a fundamental role in robotics, augmented/virtual reality (AR/VR), and autonomous driving, where understanding spatial layouts and maintaining accurate localization are crucial. In recent years, the focus has expanded beyond purely geometric reconstruction, as spatial AI~\cite{ davison2018futuremapping} increasingly emphasizes semantic understanding of 3D environments to enable intelligent behavior, meaningful interaction, and effective decision-making.
3D Gaussian Splatting (3DGS)~\cite{kerbl3Dgaussians} has emerged as a powerful representation for fast and high-fidelity 3D rendering, enabling differentiable rasterization for efficient optimization. This has led to a series of 3DGS-based SLAM systems that achieve real-time tracking and dense reconstruction~\cite{keetha2024splatam, ha2024rgbd, yan2024gs, matsuki2024gaussian, hhuang2024photoslam, wen2025scaffold, sandstrom2025splat}.

Despite these advances, many 3DGS-based SLAM methods still rely on depth sensors to obtain high-quality reconstruction~\cite{keetha2024splatam,ha2024rgbd,yan2024gs}. This reliance limits deployment on lightweight or low-cost platforms where only RGB cameras are available. While monocular alternatives have been explored, they often struggle to match the geometric accuracy and completeness of RGB-D systems~\cite{matsuki2024gaussian,hhuang2024photoslam,wen2025scaffold, sandstrom2025splat}, leaving a persistent performance gap in practical monocular settings.

Meanwhile, there is growing interest in enriching 3DGS-based SLAM with semantic understanding beyond geometry. Recent methods~\cite{li2024sgs,li2024hier} have explored semantic mapping, yet most remain limited to closed-set recognition~\cite{li2024sgs} with fixed taxonomies and curated data, restricting generalization to novel categories. Moreover, many approaches depend on dense depth input for accurate reconstruction and semantic integration~\cite{li2024hier,yang2025opengs}, which further constrains applicability in monocular scenarios.

In this work, we present a novel \textit{monocular} SLAM framework that unifies \textit{3D Gaussian Splatting} and \textit{open-set semantic understanding}, named \textit{OpenMonoGS-SLAM}. Building on the efficiency of 3DGS, our framework enables accurate and fast 3D reconstruction and camera tracking from monocular RGB input. We further move beyond closed-set limitations by integrating open-vocabulary semantic reasoning. To the best of our knowledge, this combination has not been realized in existing SLAM systems.

To achieve our goal, we leverage recent Visual Foundation Models (VFMs)~\cite{kirillov2023segment,radford2021learning,leroy2024grounding} with strong zero-shot generalization. Specifically, we utilize MASt3R-derived features~\cite{leroy2024grounding} for camera tracking and 3D point reconstruction, and employ SAM~\cite{kirillov2023segment} to produce 2D segmentation maps that are lifted into 3D with multi-view consistency for stable semantic mapping. We additionally incorporate CLIP~\cite{radford2021learning} to enable high-level open-vocabulary semantics, embedding these VFM-derived semantic cues into the 3D Gaussian field.

To better handle the inherent multi-scale nature of semantic entities, we convert SAM masks of varying sizes into a scale-aware supervision signal that reduces ambiguity from overlapping masks and encourages semantics that smoothly transition from coarse groupings to fine-grained separation.

To improve efficiency, we maintain a compact Gaussian feature map and retrieve high-dimensional CLIP semantics from a memory bank via attention-based readout, enabling rich semantic fusion without inflating per-Gaussian feature dimensionality. Finally, we combine multiple self-supervised learning objectives to learn semantic information without requiring any 3D semantic ground truth, improving both segmentation and mapping quality.

Our key contributions are summarized as follows:
\begin{itemize}
    \item We propose a monocular 3D open-set semantic SLAM framework that integrates VFM-derived semantic features into 3DGS representation. By expressing these features in a multi-scale manner, our approach enables scale-aware and open-set semantic mapping without requiring expensive depth or semantic annotation.
    \item We design an efficient semantic fusion strategy that maintains a low-dimensional Gaussian semantic map and uses attention-based readout from a CLIP feature memory bank to retrieve high-dimensional semantics. This structure allows efficient semantic retrieval and fusion during mapping, while self-supervised losses enable open-set training without dense annotations.
    \item We show that semantics and geometry mutually reinforce each other in monocular 3DGS SLAM, where semantic consistency improves map stability and geometry accuracy, while better geometry enables reliable semantic lifting.
    \item We validate our approach across standard benchmarks, showing that our method achieves high-quality semantic consistency and photorealistic rendering, comparable or superior to prior SLAM systems that rely on more restrictive inputs.
\end{itemize}

\section{RELATED WORK}
\label{sec:related_work}

\subsection{Visual Foundation Model}
Recent advances in visual foundation models have significantly impacted a broad range of computer vision tasks by enabling strong generalization and transferability. These models are typically pretrained on large-scale image datasets using self-supervised or weakly supervised objectives, and are later adapted to a variety of downstream applications. Prominent 2D vision foundation models such as DINO~\cite{caron2021dino} and CroCo~\cite{weinzaepfel2022croco}, learn dense visual representations through contrastive learning or masked image modeling. Trained without explicit human supervision, these models capture rich semantic and geometric information and have been successfully applied to a wide range of tasks, including segmentation~\cite{kirillov2023segment}, dense matching~\cite{jiang2024omniglue, edstedt2024roma}, and 3D reconstruction~\cite{leroy2024grounding, wang2025vggt}. Their dense features serve as general-purpose image descriptors, often used as backbones in more specialized frameworks.


This line of work suggests that dense correspondence, promptable segmentation, and language-aligned representations can provide strong priors for building semantic 3D understanding with minimal manual supervision. However, directly using VFM outputs in an incremental mapping setting remains challenging, since the signals can be view-dependent, multi-scale, and temporally inconsistent, requiring careful integration across frames and viewpoints.


\subsection{Visual SLAM}
Visual SLAM aims to estimate camera motion while incrementally building a consistent map from streaming images, enabling real-time localization and mapping in robotics and AR/VR~\cite{davison2007monoslam, mur2015orb}. 
Most classical systems produce sparse or semi-dense maps that are effective for tracking, yet limited for high-fidelity scene modeling and rich scene understanding~\cite{mur2015orb, engel2014lsd}.

In parallel, the mapping component has benefited from advances in dense multi-view reconstruction, ranging from Structure-from-Motion and Multi-View Stereo~\cite{schoenberger2016sfm, schoenberger2016mvs} to neural implicit and explicit representations~\cite{mildenhall2021nerf, kerbl3Dgaussians}. 
While these methods provide detailed geometry and appearance, they often rely on offline optimization, which limits their direct use in real-time SLAM.



To address this, recent work has explored integrating neural representations into incremental SLAM pipelines~\cite{keetha2024splatam, ha2024rgbd, yan2024gs, matsuki2024gaussian, hhuang2024photoslam, wen2025scaffold, sandstrom2025splat}, enabling online reconstruction while retaining high-quality geometry and appearance. Despite this progress, incorporating scalable semantics remains difficult. Many approaches focus primarily on geometry, and semantic extensions are often constrained by depth dependence or fixed label spaces, which limits adaptability to novel categories and open-set environments. Our method addresses these limitations by leveraging visual foundation models for open-set semantic SLAM, unifying promptable semantic segmentation and dense matching representations within a single framework.


\begin{figure*}[!t]
    \centering
    \vspace*{3mm}
    \includegraphics[width=0.85\textwidth]{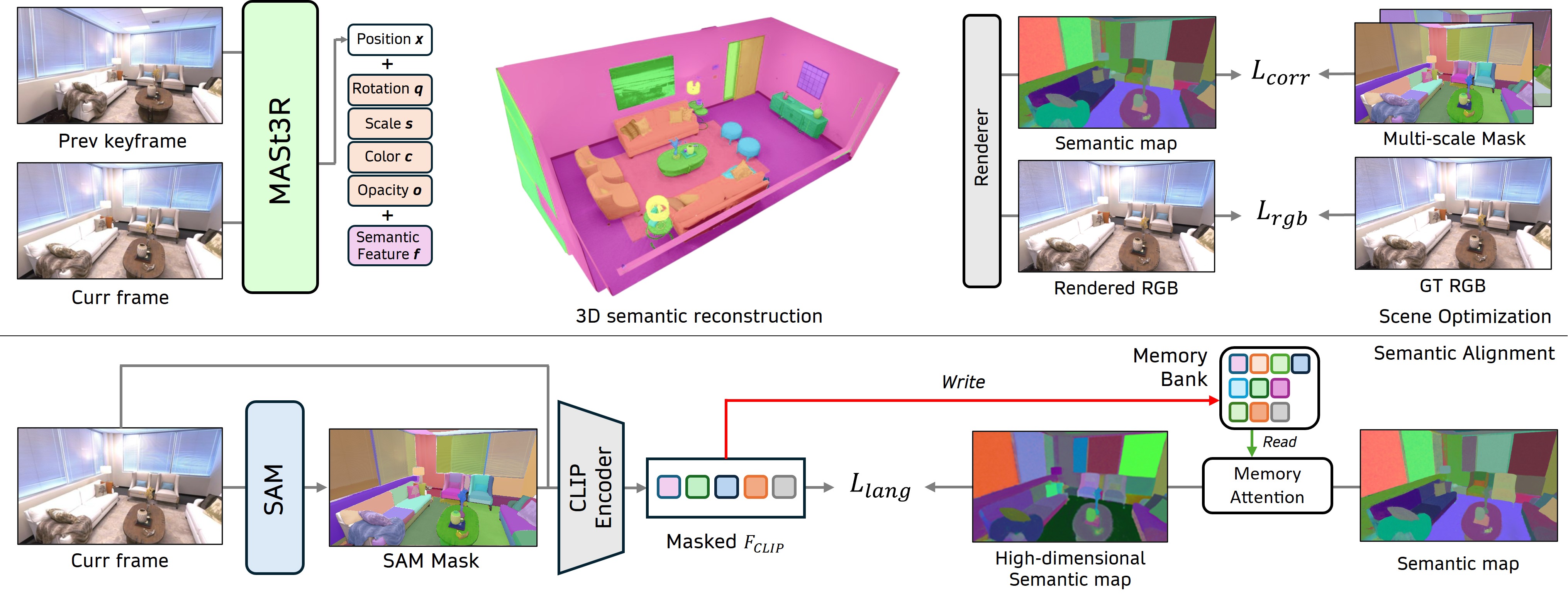}
    \caption{\textbf{Overview of Our Method.} \textbf{Top}: Given the previous keyframe and the current frame, MASt3R estimates a point map. The 3D Gaussian scene is initialized by assigning each point with Gaussian attributes and a learnable semantic feature vector. Differentiable rendering produces an RGB image and a semantic feature map, supervised by the ground-truth RGB image and multi-scale SAM masks, respectively. \textbf{Bottom}: When the current frame becomes a new keyframe, SAM predicts instance masks and masked CLIP features are extracted from the RGB image. These features update the memory bank online and provide targets for the language-guided loss. Memory attention then retrieves high-dimensional semantics to refine the Gaussian semantic map.}    
    \label{fig:overview}
    \vspace*{-5mm}
\end{figure*}

\section{METHOD}
\label{sec:method}

In this section, we detail the framework and training strategy of our proposed OpenMonoGS-SLAM. Our system builds upon MASt3R-SLAM, and incorporates semantic knowledge from Visual Foundation Models (VFMs) into a 3D Gaussian Splatting (3DGS) representation for expressive and scalable scene understanding. The overall pipeline is illustrated in Fig.~\ref{fig:overview}.

\subsection{3DGS-based Monocular SLAM} 

\subsubsection{MASt3R-SLAM} 
We adopt MASt3R-SLAM~\cite{murai2024_mast3rslam} as our backbone model for camera tracking and pointmap estimation. MASt3R employs dense per-pixel features, enabling robust tracking under challenging conditions.
We can obtain reliable correspondences by iteratively optimizing the ray errors. Given the output pointmaps $\mathbf{X}_i^i, \mathbf{X}_i^j \in \mathbb{R}^{HW \times 3}$, the optimal pixel coordinate in the reference frame $i$, for each target point $\mathbf{x} \in \mathbf{X}_i^j$ can be found with the following optimization objective,
\begin{equation}
\mathbf{p}^* = \arg\min_{\mathbf{p}} \left\| \psi\left([\mathbf{X}_i^i]_\mathbf{p}\right) - \psi(\mathbf{x}) \right\|^2,
\end{equation}
where $\mathbf{X}_i^j \in \mathbb{R}^{H \times W \times 3}$ denotes the output pointmap of image $j$ in the camera $i$'s coordinate frame, $\psi\left([\mathbf{X}_i^i]_\mathbf{p}\right), \psi(\mathbf{x})$ are a queried ray ($i$'s frame) and the target ray ($j$'s frame), normalized to unit norm, respectively. Based on the established correspondences, the camera tracking is achieved by minimizing the following ray error loss:
\begin{equation}
E_r = \sum_{m,n \in \mathbf{m}_{f,k}} \left\| \frac{\psi\left(\tilde{\mathbf{X}}_{k,n}^k\right) - \psi\left(\mathbf{T}_{kf} \mathbf{X}_{f,m}^f\right)}{w(\mathbf{q}_{m,n}, \sigma_r^2)} \right\|_\rho,
\end{equation}
\noindent
where $\tilde{\mathbf{X}}_{k,n}^k, \mathbf{X}_{f,m}^f \in \mathbb{R}^{3}$ denote the 3D coordinates of matching features between a frame $f$ and a keyframe $k$, respectively. $\mathbf{T}_{kf}$ represents the relative transformation from frame $f$ to keyframe $k$. $w(\mathbf{q}_{m,n}, \sigma_r^2)$ is a confidence-based weighting function where $\mathbf{q}_{m,n} = \sqrt{\mathbf{Q}_{f,m} \mathbf{Q}_{f,n}^k}$ is the match confidence score. The Huber norm $\| \cdot \|_\rho$ is applied to improve robustness against outliers. 

\subsubsection{3D Gaussian Representation} 
The camera poses estimated through this tracking procedure are used to project MASt3R-generated 3D points into a global coordinate frame. We then initialize a set of 3D Gaussians at these positions. Each Gaussian maintains color attributes as well as learnable semantic features. 
\begin{equation}
\mathcal{G}_i = \{\mathbf{x}_i, \mathbf{q}_i, \mathbf{s}_i, \alpha_i, \mathbf{c}_i, \mathbf{f}_i\},
\end{equation}
where $\mathbf{x}_i$ denotes the 3D position, $\mathbf{q}_i$ the quaternion orientation, $\mathbf{s}_i$ the scale, $\alpha_i$ the opacity, $\mathbf{c}_i$ the RGB color, and $\mathbf{f}_i \in \mathbb{R}^d$ the learnable semantic feature.

Since Gaussians are initialized using a pixel-aligned 3D pointmap from MASt3R, they can be accurately placed in 3D space, enabling high-quality rendering without the need for additional densification.
Once initialized, the rendering of RGB and semantic features for each pixel coordinate $\mathbf{p}$ is performed via depth-ordered alpha blending as follows,
\begin{equation}
\mathbf{C}_\mathbf{p} = \sum_{i=1}^N \mathbf{c}_i \alpha_i \prod_{j=1}^{i-1}(1 - \alpha_j),
\end{equation}
\begin{equation}
\mathbf{F}_\mathbf{p} = \sum_{i=1}^N \mathbf{f}_i \alpha_i \prod_{j=1}^{i-1}(1 - \alpha_j).
\end{equation}

Similar to \cite{ha2024rgbd}, we supervise 3D Gaussians not only with keyframes but also with additional mapping frames, which are used exclusively for optimization. This is necessary since the keyframes selected by MASt3R-SLAM are too sparse to ensure consistent updates across all regions. This representation yields a compact and expressive structure capable of both photorealistic rendering and semantic reasoning.

\subsection{Multi-Scale Semantic Learning}
\label{sec:ms_semantic}

Recent works have demonstrated the importance of modeling the multi-scale nature of semantic objects for effective 3D scene understanding~\cite{kerr2023lerf, li2024hier, cen2025segment}. Building on this insight, we aim to guide 3D learning using 2D segmentation outputs from VFMs. To improve generalization to various object sizes and avoid dependence on fixed semantic taxonomies, we adopt a scale-aware strategy. We leverage SAM to produce 2D object masks from each input image. These masks vary in size, reflecting the multi-scale nature of the objects. These 2D masks are then lifted into 3D using depth estimates from MASt3R-SLAM and the known camera intrinsics.

To train scale-aware semantic features, we convert the set of SAM masks into a scale-conditioned supervision signal using $S$ predefined scale levels, inspired by scale-aware contrastive distillation~\cite{cen2025segment}. Concretely, each lifted mask is assigned a 3D scale (e.g., its 3D spatial extent after lifting), and we discretize the scale axis into $S$ levels $\{s\}$ to construct supervision that smoothly transitions from coarse object grouping to fine part-level separation.

For each level $s$, we define a binary pixel identity vector $\mathbf{V}(s,\mathbf{p})\in\{0,1\}^{N}$, where $N$ is the number of lifted masks, indicating which masks are active for pixel $\mathbf{p}$ at that scale. Let $\{\mathbf{M}_i\}_{i=1}^{N}$ denote the lifted masks, ordered by their 3D scales from coarse to fine. A key challenge is multi-granularity ambiguity: a pixel may belong to multiple overlapping masks that represent different granularities (e.g., part vs. whole object). To resolve such overlaps in a scale-dependent manner, at scale $s$ we activate mask $\mathbf{M}_i$ for pixel $\mathbf{p}$ only if $\mathbf{p}\in \mathbf{M}_i$ and there exists no smaller mask $\mathbf{M}_j$ such that $\mathbf{p}\in \mathbf{M}_j$ and
$s \leq \text{size}(\mathbf{M}_j) < \text{size}(\mathbf{M}_i)$.

This rule can be interpreted as follows: when $s$ is small (fine levels), many smaller masks satisfy the condition and thus suppress larger overlapping masks, assigning ambiguous regions primarily to fine masks. In contrast, when $s$ is large (coarse levels), fewer smaller masks remain eligible, so larger masks stay active more often, yielding broader, coarse-level groupings. As a result, $\mathbf{V}(s,\mathbf{p})$ provides scale-conditioned membership that reduces ambiguity across granularities.

Finally, we convert the scale-conditioned identity vectors into a binary \emph{mask correspondence} signal, which is used in our multi-view contrastive objective (Sec.~\ref{sec:ss_objectives}). 
Specifically, we define the mask correspondence between two pixels $\mathbf{p}_1$ and $\mathbf{p}_2$ at scale $s$ as
\begin{equation}
\mathrm{Corr}_m(s,\mathbf{p}_1,\mathbf{p}_2)=\ind\!\left( \mathbf{V}(s,\mathbf{p}_1)^\top \mathbf{V}(s,\mathbf{p}_2) > 0 \right),
\end{equation}
where $\ind(\cdot)$ denotes the indicator function that returns $1$ if the condition holds and $0$ otherwise.

\subsection{Language-embedded Semantic Memory Bank}

While SAM provides instance-level segmentation, these masks lack semantic meaning without language grounding. \cite{kerr2023lerf} has demonstrated the effectiveness of embedding natural language into radiance fields for semantic understanding. Inspired by this, we produce embeddings for each segmented region using CLIP features to inject language priors into our 3D representation. However, storing raw CLIP embeddings for all Gaussians is prohibitively memory-intensive.

To address this, similar to M3~\cite{zou20253d}, we maintain a compact memory bank that stores representative CLIP embeddings to cover the semantic space efficiently. While M3 proposes this mechanism in an offline setting, directly applying it to SLAM is impractical due to the need for online updates. To adapt this idea, we introduce an online mechanism: when a new keyframe is selected, we compute cosine similarity between the current CLIP embedding and the memory bank entries, which store representative CLIP embeddings from previous keyframes. If the similarity is lower than $\tau_m$, which means sufficiently different, we append it to the memory bank. This ensures a diverse but compact coverage of the semantic space. 

During mapping, the rendered 2D semantic features serve as queries and attend to the memory bank via an attention mechanism to retrieve high-dimensional semantic features. Given a rendered feature $\mathbf{F}_\mathbf{p} \in \mathbb{R}^d$, the detailed formulation is as follows:
\begin{equation}
\hat{\mathbf{F}}_\mathbf{p} = \mathrm{softmax}((\mathbf{W}_{proj}\mathbf{F}_\mathbf{p}) \mathcal{M}^\top) \mathcal{M},
\end{equation}
where $\mathcal{M} \in \mathbb{R}^{M \times D}$ is the memory bank composed of $M$ high-dimensional CLIP features, and $\mathbf{W}_{proj} \in \mathbb{R}^{D \times d}$ is a linear projection layer that aligns the query space with the memory space, and $\mathrm{softmax}:\mathbb{R}^M \rightarrow [0,1]^M$ returns attentional scores. The output $\hat{\mathbf{F}}_\mathbf{p} \in \mathbb{R}^D$ contains the attended high-dimensional features for each Gaussian.


This allows us to enrich each Gaussian with semantic information while maintaining a compact and efficient Gaussian map. At the same time, we fully leverage high-dimensional language features to maximize semantic expressiveness.


\subsection{Self-Supervised Learning Objectives}
\label{sec:ss_objectives}

We employ a combination of self-supervised losses to jointly optimize geometry, appearance, and semantics. This synergy between losses allows our model to build expressive 3D maps that are photorealistic and semantically meaningful.

\textit{a) Photometric Supervision:} We render novel views from the current 3D Gaussian representation and minimize a weighted combination of $\mathcal{L}_1$ and the structural similarity index loss $\mathcal{L}_{\text{SSIM}}$ between the rendered and ground truth images. This encourages the learned Gaussians to maintain accurate spatial distribution and photorealistic visual fidelity.
\begin{equation}
\mathcal{L}_{\text{rgb}} = (1 - \lambda) \mathcal{L}_1 + \lambda (1 - \mathcal{L}_{\text{SSIM}}).
\end{equation}

\textit{b) Multi-View Semantic Consistency:} To enforce semantic consistency across different views, we apply a multi-view contrastive learning objective that pulls semantically similar Gaussians closer and pushes dissimilar ones apart in the feature space. Following SAGA~\cite{cen2025segment}, the corresponding contrastive loss between two sampled pixel coordinates $\mathbf{p_1},\mathbf{p_2}$ is written as,
\begin{align}
\mathcal{L}_{\text{corr}} = \frac{1}{S|\mathcal{P}|^2}\sum_{s=1}^S \sum_{\mathbf{p}_1 \in \mathcal{P}} \sum_{\mathbf{p}_2 \in \mathcal{P}} (1 - 2 \cdot \mathrm{Corr}_m(s, \mathbf{p}_1, \mathbf{p}_2)) \nonumber \\
\cdot {\max(\mathrm{Corr}_f(s, \mathbf{p}_1, \mathbf{p}_2), 0)},
\end{align}
where $s$ denotes the scale index, and $\mathrm{Corr}_m, \mathrm{Corr}_f$ represent mask and feature correspondence, respectively. 
We refer to Sec.~\ref{sec:ms_semantic} for the definition of $\mathrm{Corr}_m$. 
We compute $\mathrm{Corr}_f$ as the cosine similarity of rendered semantic features:
\begin{equation}
\mathrm{Corr}_f(s,\mathbf{p}_1,\mathbf{p}_2)=\left\langle \mathbf{F}_{\mathbf{p}_1}, \mathbf{F}_{\mathbf{p}_2}\right\rangle .
\end{equation}
We sample $|\mathcal{P}|=2,000$ points and use $S=4$ scales for multi-scale training.
By encouraging consistent semantic features across views, this loss also regularizes the underlying 3D Gaussian map, which empirically leads to more stable camera pose estimation.








\textit{c) Language-Guided Semantic Alignment:} To integrate open-set semantic priors from VFMs, we supervise the learned features with CLIP embeddings via a regression objective:

\begin{equation}
\mathcal{L}_{\text{lang}} = 
\frac{1}{|\mathcal{P}|} \sum_{\mathbf{p} \in \mathcal{P}} 
\left[ 
\left(1 \!-\! \cos\!\left(\hat{\mathbf{F}}_{\mathbf{p}}, \mathbf{F}_{\mathbf{p}}^{\text{CLIP}}\right)\!\right) 
\!+\! 
\left\| \hat{\mathbf{F}}_{\mathbf{p}} - \mathbf{F}_{\mathbf{p}}^{\text{CLIP}} \right\|_2^2 
\right].
\end{equation}

\textit{d) Final Loss:} We combine all losses into a unified training objective:
\begin{equation}
\mathcal{L}_{\text{total}} = \lambda_{\text{rgb}} \mathcal{L}_{\text{rgb}} + \lambda_{\text{corr}} 
\mathcal{L}_{\text{corr}} + \lambda_{\text{lang}} \mathcal{L}_{\text{lang}}.
\end{equation}
By leveraging these complementary loss terms in a self-supervised manner, our model effectively balances geometric accuracy, photometric fidelity, semantic consistency, and language grounding. This enables OpenMonoGS-SLAM to construct robust and interpretable 3D maps from monocular inputs, even in challenging open-set environments.


\section{EXPERIMENTS}

\begin{table*}[htbp]
\centering
\vspace*{3mm}
\resizebox{0.8\textwidth}{!}{%
\begin{tabular}{lcccccccccccc}
\toprule
\textbf{Methods} & \textbf{Input} & \textbf{Metrics} & \textbf{Avg.} & \textbf{room0} & \textbf{room1} & \textbf{room2} & \textbf{office0} & \textbf{office1} & \textbf{office2} & \textbf{office3} & \textbf{office4} \\
\midrule
\multicolumn{12}{l}{\textbf{Visual SLAM}} \\
\multirow{3}{*}{MonoGS \cite{matsuki2024gaussian}} & \multirow{3}{*}{RGB} & PSNR ↑ & 27.77 & 26.23 & 25.28 & 27.87 & 31.28 & 33.61 & 23.78 & 27.97 & 26.15 &  \\
& & SSIM ↑ & 0.858 & 0.821 & 0.776 & 0.869 & 0.886 & 0.916 & 0.830 & 0.891 & 0.878 \\
& & LPIPS ↓ & 0.203 & 0.179 & 0.304 & 0.184 & 0.181 & 0.142 & 0.262 & 0.136 & 0.237 \\ \hline
\multirow{3}{*}{Photo-SLAM \cite{hhuang2024photoslam}} & \multirow{3}{*}{RGB} & PSNR ↑ & 30.37 & 29.73 & 26.65 & 31.97 & 35.27 & 28.74 & 30.72 & 29.25 & \second{30.62} \\
& & SSIM ↑ & 0.904 & 0.874 & \second{0.828} & 0.930 & 0.941 & 0.878 & 0.937 & 0.912 & \second{0.931} \\
& & LPIPS ↓ & 0.161 & 0.156 & 0.246 & 0.115 & 0.125 & 0.227 & 0.143 & 0.139 & \second{0.138} \\ \hline
\multirow{3}{*}{SEGS-SLAM \cite{wen2025scaffold}} & \multirow{3}{*}{RGB} & PSNR ↑ & \second{33.54} & \second{31.25} & \second{27.27} & \second{35.09} & \best{38.56} & \best{38.64} & \best{34.42} & \best{33.90} & 29.35 \\
& & SSIM ↑ & \second{0.927} & \second{0.901} & 0.826 & \second{0.953} & \second{0.969} & \second{0.957} & \second{0.955} & \second{0.952} & 0.908 \\
& & LPIPS ↓ & \second{0.104} & \second{0.092} & \second{0.203} & \best{0.056} & \best{0.064} & \second{0.108} & \best{0.082} & \second{0.074} & 0.153 \\ \hline
\multirow{3}{*}{\textbf{Ours (SAM1.0)}} & \multirow{3}{*}{RGB} & PSNR ↑ & \best{34.47} & \best{33.53} & \best{30.64} & \best{35.16} & \second{38.41} & \second{38.59} & \second{32.10} & \second{33.49} & \best{33.86} \\
& & SSIM ↑ & \best{0.957} & \best{0.953} & \best{0.914} & \best{0.967} & \best{0.969} & \best{0.971} & \best{0.959} & \best{0.959} & \best{0.963} \\
& & LPIPS ↓ & \best{0.086} & \best{0.066} & \best{0.153} & \second{0.075} & \second{0.065} & \best{0.084} & \second{0.085} & \best{0.067} & \best{0.093} \\
 
\midrule
\multicolumn{12}{l}{\textbf{Semantic SLAM}} \\
\multirow{3}{*}{DNS-SLAM \cite{li2024dns}} & \multirow{3}{*}{RGB-D} & PSNR ↑ & 22.89 & 22.45 & 24.61 & 25.27 & 24.09 & 25.28 & 21.39 & 21.87 & 18.20 \\
& & SSIM ↑ & 0.851 & 0.844 & 0.882 & 0.900 & 0.891 & 0.809 & 0.880 & 0.870 & 0.832 \\
& & LPIPS ↓ & 0.231 & 0.231 & 0.230 & 0.231 & 0.195 & 0.259 & 0.251 & 0.256 & 0.283 \\ \hline
\multirow{3}{*}{SGS-SLAM \cite{li2024sgs}} & \multirow{3}{*}{RGB-D} & PSNR ↑ & 32.11 & 29.83 & \best{31.60} & 32.68 & 36.75 & 37.28 & 29.72 & 28.63 & 30.36 \\
& & SSIM ↑ & 0.922 & 0.891 & \second{0.915} & 0.941 & 0.959 & 0.949 & 0.916 & 0.897 & 0.907 \\
& & LPIPS ↓ & 0.343 & 0.153 & 0.168 & 0.153 & 0.136 & 0.185 & 0.176 & 0.186 & 0.213 \\  \hline
\multirow{3}{*}{Hier-SLAM \cite{li2024hier}} & \multirow{3}{*}{RGB-D} & PSNR ↑ & \second{33.57} & \second{30.16} & 31.15 & \second{33.46} & \second{37.36} & \second{39.12} & \second{31.27} & \second{30.83} & \second{33.19} \\
& & SSIM ↑ & \second{0.933} & \second{0.896} & 0.902 & \second{0.947} & \second{0.961} & \second{0.965} & \second{0.932} & \second{0.925} & \second{0.936} \\
& & LPIPS ↓ &  \best{0.061} & \second{0.085} &  \best{0.077} &  \best{0.048} &  \best{0.041} &  \best{0.034} &  \best{0.066} &  \best{0.060} &  \best{0.079} \\  \hline
\multirow{3}{*}{\textbf{Ours (GT)}} & \multirow{3}{*}{RGB} & PSNR ↑ & \best{35.12} & \best{33.29} & \second{31.20} & \best{36.27} & \best{38.68} & \best{39.98} & \best{32.80} & \best{34.42} & \best{34.33} \\
& & SSIM ↑ & \best{0.960} & \best{0.952} & \best{0.918} & \best{0.972} & \best{0.970} & \best{0.976} & \best{0.963} & \best{0.966} & \best{0.966} \\
& & LPIPS ↓ & \second{0.080} &  \best{0.067} & \second{0.151} & \second{0.066} & \second{0.064} & \second{0.073} & \second{0.081} &  \best{0.058} & \second{0.084} \\
\bottomrule
\end{tabular}%
}
\caption{Quantitative comparisons on the Replica dataset. Best results are highlighted as \colorbox{mygreen}{\textbf{FIRST}}, \colorbox{myyellow}{SECOND}. In the semantic SLAM setting, all baselines utilize RGB-D input, whereas our method relies solely on RGB. Despite this, it achieves comparable or better performance, highlighting the effectiveness of our approach.}
\label{tab:mapping_result}
\vspace{-3mm}
\end{table*}

\begin{figure*}[!h]
    \centering
    \includegraphics[width=0.8\textwidth]{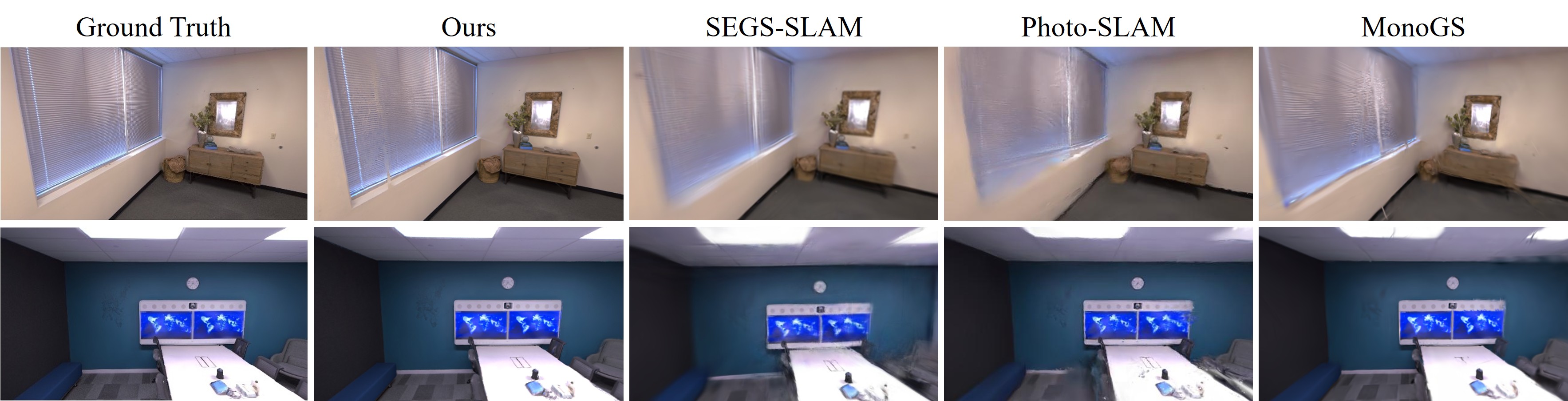}
    \caption{Qualitative comparisons of novel view synthesis on the Replica dataset.}    
    \label{fig:main}
    \vspace*{-5mm}
\end{figure*}

\subsection{Experiments Setup}
\textbf{Implementation Details.}
Our OpenMonoGS-SLAM system is built upon the MASt3R-SLAM framework. We initialize the 3D Gaussian attributes from dense 3D point maps generated from the pretrained MASt3R and adopt the gsplat framework~\cite{ye2025gsplat} for efficient and scalable differentiable rendering of the 3D scene.

For multi-scale learning, we use $S=4$ scale levels and the memory threshold is set to $\tau_m = 0.9$. The loss weight parameters are $\lambda_{\text{photo}} = 1.0$, $\lambda_{\text{corr}} = 0.05$, and $\lambda_{\text{lang}} = 0.05$. For visual foundation models (VFMs), we utilize MASt3R, SAM (ViT-H), and CLIP (ViT-B) with their default configurations. All experiments are conducted on a single NVIDIA RTX 4090 GPU with 24GB VRAM. Mapping is performed for 30K iterations, adding a mapping frame every 10 frames in periods where no keyframe is selected. Following MASt3R-SLAM's preprocessing pipeline, we resize all input images to a width of 512 pixels for both our method and the baselines to ensure a fair comparison.


To address open-set semantic segmentation, we incorporate Grounded-SAM~\cite{ren2024grounded}, which enables segmentation via text prompt guidance without requiring predefined class labels. For fair comparison with baselines, all offline models are trained using a consistent setting, selecting every 10th frame as a training view and optimizing for 30K iterations.

\noindent\textbf{Dataset and Evaluation Metrics.}
We evaluate our method on the Replica~\cite{straub2019replica} dataset, which contains ground-truth semantic labels provided by \cite{zhi2021place}. We also evaluate on the TUM RGB-D~\cite{tum} and ScanNet~\cite{dai2017scannet} benchmark under a monocular setting, reporting tracking and reconstruction quality. For camera tracking evaluation, we compute the absolute trajectory error (ATE) using the root mean square error (RMSE) metric 
To assess mapping quality, we evaluate photometric reconstruction using Peak Signal-to-Noise Ratio (PSNR), Structural Similarity Index Measure~\cite{wang2004ssim} (SSIM), and Learned Perceptual Image Patch Similarity~\cite{zhang2018lpips} (LPIPS). Semantic segmentation performance is quantified using mean Intersection over Union (mIoU). Frequency Weight IoU (FWIoU) and Accuracy (Acc) are used for ablation studies.


\noindent\textbf{Baselines.}
We evaluate tracking and novel view synthesis performance by comparing our method with representative monocular visual SLAM~\cite{matsuki2024gaussian, hhuang2024photoslam, wen2025scaffold, sandstrom2025splat} and RGB-D semantic SLAM~\cite{li2024dns, li2024sgs, li2024hier} baselines. For the evaluation of open-set segmentation, we evaluate our performance with offline feature-based segmentation methods~\cite{zhou2024feature, ye2024gaussian}. 
All baselines are implemented using their official code repositories. To ensure fairness, we fix the number of mapping iterations to 30k for all methods. If a baseline is originally trained with fewer iterations, we fine-tune it until it reaches 30k iterations.

\subsection{Results}

\textbf{Novel View Synthesis.} We evaluate novel view rendering using all frames except the keyframes used for training. As shown in Tab.~\ref{tab:mapping_result} and Fig.~\ref{fig:main}, our method consistently outperforms both visual and semantic SLAM baselines in most scenes, particularly in SSIM, indicating superior preservation of structural information. This improvement stems from jointly learning semantics and appearance; incorporating semantic cues helps maintain object-level consistency and preserve structural boundaries, leading to more accurate and coherent reconstructions. For fair evaluation in semantic SLAM comparisons, our model is trained using ground-truth semantic masks instead of SAM-based masks. Note that while semantic SLAM baselines utilize ground-truth depth as additional input, our method relies solely on RGB yet still achieves superior performance.
We additionally report results on two real-world benchmarks. On TUM-D (Tab.~\ref{tab:tum_quantitative}), our method achieves the highest PSNR and SSIM and the second-best LPIPS, demonstrating strong generalization beyond Replica. On ScanNet (Tab.~\ref{tab:scannet_quantitative}), our method attains the best rendering quality across all metrics against monocular baselines. Moreover, despite using SAM-based masks rather than ground-truth annotations, it outperforms all semantic SLAM baselines across all metrics.



\begin{table}[!ht]
\centering
\renewcommand{\arraystretch}{1.2} 
\resizebox{0.9\linewidth}{!}{
\begin{tabular}{l|l|cccc} 
\hline
\textbf{Category} & \textbf{Methods} & \textbf{ATE}$\downarrow$ &
\textbf{PSNR}$\uparrow$ & \textbf{SSIM}$\uparrow$ &
\textbf{LPIPS}$\downarrow$ \\
\hline
\multirow{3}{*}{Visual SLAM} &
MonoGS~\cite{matsuki2024gaussian} & 4.40 & 20.88 & 0.694 & 0.358 \\
& Photo-SLAM~\cite{hhuang2024photoslam} & 1.80 & 19.28 & 0.673 & 0.334 \\
& SEGS-SLAM~\cite{wen2025scaffold} & \best{1.14} & \second{23.20} &
\second{0.769} & \best{0.271} \\
\hline
Semantic SLAM & \textbf{Ours (SAM1.0)} &
\second{1.44} & \best{23.33} & \best{0.813} & \second{0.284} \\
\hline
\end{tabular}
}
\caption{Quantitative comparisons on the TUM-D dataset.}
\label{tab:tum_quantitative}
\vspace{-5mm}
\end{table}

\begin{table}[!ht]
\centering
\renewcommand{\arraystretch}{1.2} 
\resizebox{0.9\linewidth}{!}{
\begin{tabular}{l|l|cccc} 
\hline
\textbf{Category} & \textbf{Methods} & \textbf{ATE}$\downarrow$ &
\textbf{PSNR}$\uparrow$ & \textbf{SSIM}$\uparrow$ &
\textbf{LPIPS}$\downarrow$ \\
\hline
\multirow{2}{*}{Visual SLAM} &
Splat-SLAM~\cite{sandstrom2025splat} & 8.25 & \second{24.12} & 0.796 & 0.359 \\
& SEGS-SLAM~\cite{wen2025scaffold} & \second{7.66} & 23.62	& \second{0.846} & \second{0.304} \\
\hline
\multirow{3}{*}{Semantic SLAM} &
SGS-SLAM \cite{li2024sgs} & \second{9.86} & \second{22.27} & \second{0.753} & \second{0.348} \\
& Hier-SLAM \cite{li2024hier} & 10.53 & 21.78 & 0.739 & 0.355 \\
& \textbf{Ours (SAM1.0)} &
\best{5.39} & \best{26.21} & \best{0.860} & \best{0.273} \\
\hline
\end{tabular}
}
\caption{Quantitative comparisons on the ScanNet dataset.}
\label{tab:scannet_quantitative}
\vspace{-2mm}
\end{table}

\noindent\textbf{Camera Tracking.} As shown in Tab.~\ref{tab:tracking_result}, our method significantly outperforms existing monocular visual SLAM approaches in absolute trajectory accuracy. It is also robust across diverse indoor scenes, exhibiting low variance and consistently achieving low ATE RMSE. We attribute this stability to visual foundation models trained on large-scale and diverse data, which provide rich semantic and geometric priors and thus enable more reliable pose estimation under challenging conditions. On the real-world TUM-D benchmark (Tab.~\ref{tab:tum_quantitative}), our method attains the second-lowest ATE with a small gap to the best baseline, indicating competitive tracking performance under monocular input, while on ScanNet (Tab.~\ref{tab:scannet_quantitative}) it achieves the lowest ATE among all baselines, demonstrating strong robustness on real-world indoor scenes.

\begin{table}[!h]
\centering
\resizebox{\linewidth}{!}{
\renewcommand{\arraystretch}{1.2} 
\begin{tabular}{l|ccccccccc}
\hline
\textbf{Methods} & \textbf{Avg.} & \textbf{R0} & \textbf{R1} & \textbf{R2} & \textbf{Of0} & \textbf{Of1} & \textbf{Of2} & \textbf{Of3} & \textbf{Of4} \\ \hline
MonoGS~\cite{matsuki2024gaussian} & 30.48  & 12.98 & 48.22 & 12.11 & 26.14 & 19.20 & 47.30 & 8.49 & 69.43 \\
Photo-SLAM~\cite{hhuang2024photoslam} & 10.63 & \second{1.60} & \second{22.56} & 3.36 & 2.11 & 30.02 & 6.95 & \second{2.04} & \second{16.38} \\
SEGS-SLAM~\cite{wen2025scaffold} & \second{9.25} & \best{1.07} & 35.09 & \best{0.22} & \second{1.70} & \best{0.74} & \best{0.62} & \best{1.11} & 33.40 \\
\textbf{Ours (SAM 1.0)} & \best{1.60} & 1.96 & \best{1.18} & \second{1.08} & \best{1.16} & \second{0.94} & \second{1.28} & 3.32 & \best{1.45} \\ \hline
\end{tabular}}
\caption{Camera tracking results on the Replica dataset. ATE RMSE in cm is reported. Our method demonstrates robust and consistent tracking performance across all scenes, achieving the lowest average error.}
\label{tab:tracking_result}
\vspace*{-5mm}
\end{table}

\noindent\textbf{Open-set Segmentation Results.}
To evaluate open-set segmentation with text prompts, we use Grounded-SAM~\cite{ren2024grounded} to generate text-conditioned 2D masks. Since both OpenMonoGS-SLAM and Feature 3DGS~\cite{zhou2024feature} support prompt-based segmentation on rendered feature maps, we use the same evaluation text descriptions as prompts to obtain binary masks for each target concept. These masks are applied to the rendered feature maps to produce prompt-based segmentations within the visible regions, enabling a fair assessment of language-driven concept localization without relying on predefined semantic classes.

In contrast, Gaussian Grouping~\cite{ye2024gaussian} predicts semantics via a separate classifier and does not support prompt-based segmentation on rendered features. Therefore, following the official Gaussian Grouping protocol, we compute IoU between the text-prompted masks generated by Grounded-SAM and the predicted masks from Gaussian Grouping. To reduce the effect of irrelevant regions in the open-set setting, we report mean IoU only for pairs with IoU $>0.2$, matching the default threshold in their implementation.

\begin{figure}[!ht]
    \centering
    \includegraphics[width=\columnwidth]{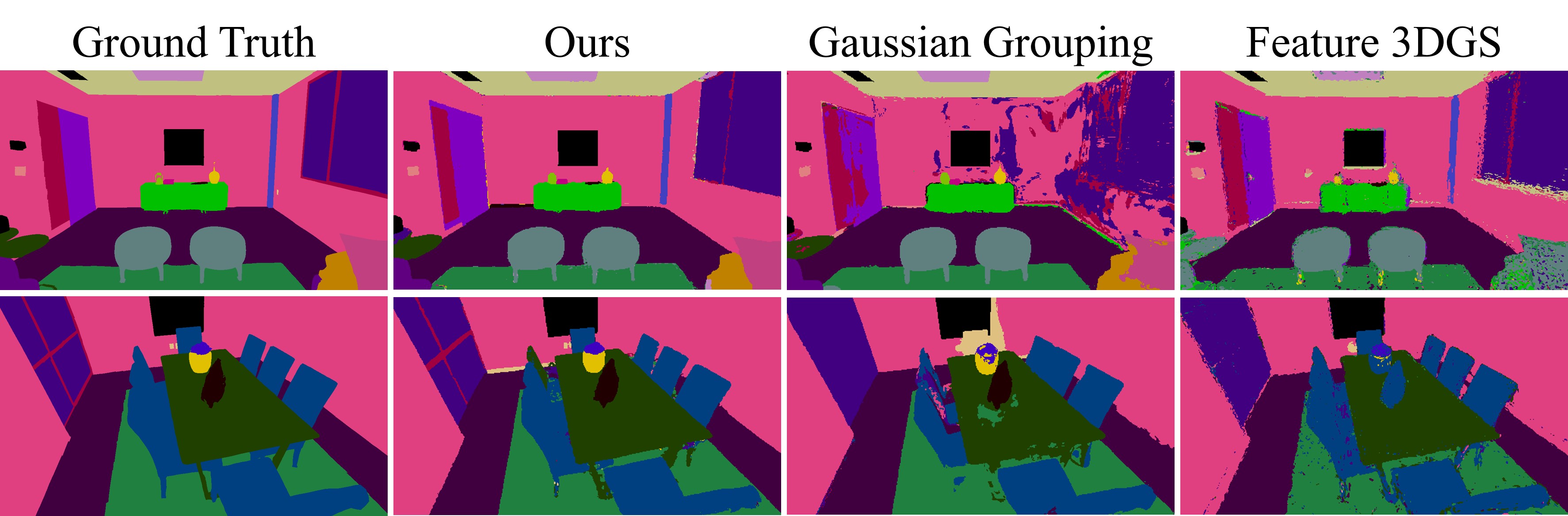}
    \caption{Qualitative comparisons of open-set segmentation on the Replica dataset. Our method produces cleaner and more complete segmentation masks, particularly for fine-grained structures.}
    \label{fig:open}
    \vspace{-3mm}
\end{figure}

As shown in Tab.~\ref{tab:Open-set segementation}, our method achieves the best open-set segmentation performance across all scenes. This improvement is largely attributed to our effective memory-based integration of CLIP language features into the Gaussian feature space, which improves alignment between rendered features and diverse textual prompts, leading to more accurate open-set segmentation (Fig.~\ref{fig:open}).



\begin{table}[!ht]
\centering
\resizebox{\linewidth}{!}{
\renewcommand{\arraystretch}{1.2} 
\begin{tabular}{l|ccccccccc}
\hline
\textbf{Methods} & \textbf{Avg.} & \textbf{R0} & \textbf{R1} & \textbf{R2} & \textbf{Of0} & \textbf{Of1} & \textbf{Of2} & \textbf{Of3} & \textbf{Of4} \\ \hline

Feature 3DGS\cite{zhou2024feature}  & 0.571   & 0.506         & 0.569       & 0.527        & 0.633            & 0.598            & 0.613            & 0.529       & 0.593     \\

Gaussian Grouping\cite{ye2024gaussian} &\second{0.690}    & \second{0.613}       & \second{0.631}       & \second{0.732}       & \second{0.744}        & \second{0.683}            & \second{0.763}            & \second{0.664}            & \second{0.686}            \\

\textbf{Ours (SAM1.0)} &  \best{0.845}  & \best{0.832} & \best{0.761} & \best{0.873} & \best{0.837} & \best{0.882} & \best{0.851} & \best{0.847} & \best{0.873} \\ \hline
\end{tabular}
    }
\caption{Quantitative comparisons of open-set segmentation on the Replica dataset. mIoU from text prompts is reported.}
\label{tab:Open-set segementation}    
\end{table}

\noindent\textbf{Closed-set Segmentation Results.}
While our primary focus is open-set semantic SLAM, we additionally evaluate closed-set segmentation using ground-truth semantic annotations to demonstrate the general applicability of our framework.

All semantic SLAM baselines directly use ground-truth semantic IDs during optimization, enabling straightforward evaluation by computing mean Intersection over Union (mIoU) between predicted and ground-truth labels. In contrast, our method is class-agnostic and does not assume explicit semantic supervision. To enable a fair closed-set comparison, we add an auxiliary cross-entropy loss term during SLAM optimization that leverages the ground-truth semantic IDs, while keeping the rest of our objectives unchanged. After optimization, we follow the same protocol as the baselines and report mIoU between the predicted and ground-truth masks for each semantic class.

As shown in Tab.~\ref{tab:Close-set segementation}, our method achieves the best performance across all scenes except R1 and Of1, where it remains highly competitive. These results indicate that, although designed for open-set scenarios, the proposed approach also effectively transfers to closed-set segmentation.

\begin{table}[!ht]
\centering
\resizebox{\linewidth}{!}{
\renewcommand{\arraystretch}{1.2} 
\begin{tabular}{l|ccccccccc}
\hline
\textbf{Methods} & \textbf{Avg.} & \textbf{R0} & \textbf{R1} & \textbf{R2} & \textbf{Of0} & \textbf{Of1} & \textbf{Of2} & \textbf{Of3} & \textbf{Of4} \\ \hline
DNS-SLAM\cite{li2024dns}  & 0.742   & 0.818       & 0.831       & 0.748       &  0.671        & 0.798          & 0.759            & 0.721            & 0.589            \\
SGS-SLAM\cite{li2024sgs}   & \second{0.866}  &  0.862  &    \best{0.887}  &  \second{0.865}  &    \second{0.886}  & \best{0.927}  & \second{0.854}            & \second{0.782}            & \second{0.866}             \\

Hier-SLAM\cite{li2024hier}  & 0.804  &  \second{0.869}  &    0.767  &  0.844  &    0.817  & 0.807  & 0.830            & 0.740            & 0.793             \\

\textbf{Ours(GT)} & \best{0.896} & \best{0.886} & \second{0.881} & \best{0.900} & \best{0.904} & \second{0.915} & \best{0.909} & \best{0.875} & \best{0.901} \\ \hline
\end{tabular}
    }
\caption{Quantitative comparisons of close-set segmentation on the Replica dataset. mIoU based on ground-truth semantic labels is reported.}
\label{tab:Close-set segementation}
\vspace{-2mm}
\end{table}

\begin{figure}[!ht]
    \centering
    \includegraphics[width=\columnwidth]{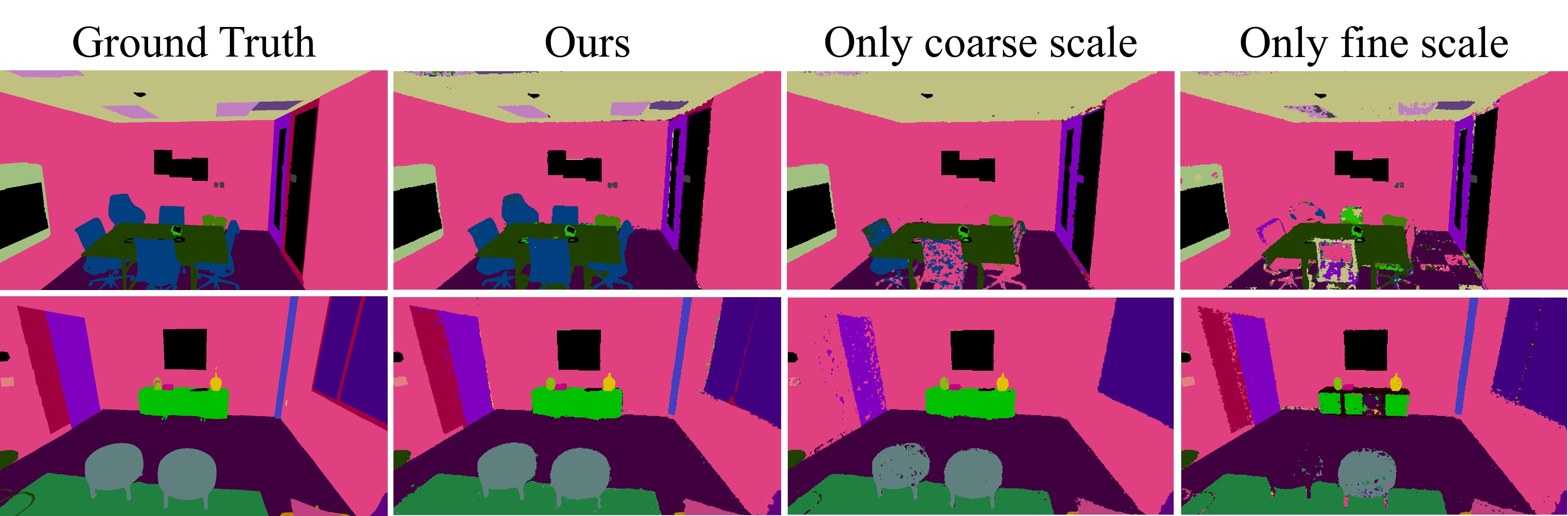}
    \caption{Qualitative comparison of multi-scale ablations on the Replica dataset. Multi-scale supervision produces more coherent masks with finer details.}  
    \label{fig:ablation_multi_scale}
    \vspace{-2mm}
\end{figure}

\begin{figure}[!ht]
    \centering
    \includegraphics[width=\columnwidth]{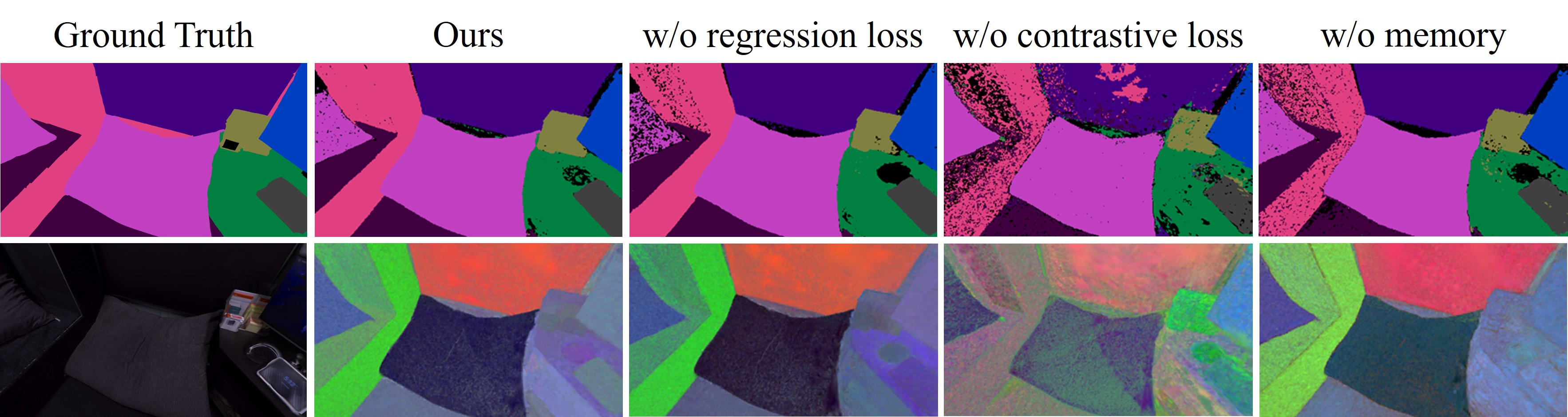}
    \caption{Qualitative comparison of ablated components on the Replica dataset. The top row shows the open-set segmentation results, while the bottom row presents the corresponding rendered features (with the ground-truth RGB image in the first column).}  
    \label{fig:ablation}
    \vspace*{-5mm}
\end{figure}

\noindent\textbf{Ablations and  Analysis.} To validate the effectiveness of each component in our framework, we conduct comprehensive ablation studies with both qualitative and quantitative evaluations as presented in Fig.~\ref{fig:ablation_multi_scale}, Fig.~\ref{fig:ablation} and Tab.~\ref{tab:ablation}. All methods are trained for 30K iterations on the full set of Replica scenes for a fair comparison. For segmentation evaluation, we use all object masks available in the semantic ground truth as prompts for each scene.


We first analyze the contribution of each loss term. Removing the multi-view contrastive loss (``w/o contrastive loss") causes a substantial mIoU drop of approximately 27\%, accompanied by clear declines in FWIoU and pixel accuracy, highlighting the importance of enforcing cross-view semantic consistency. Excluding the regression loss (``w/o regression loss") also degrades performance, indicating its essential role in guiding semantic alignment. As shown in Fig.~\ref{fig:ablation}, removing either loss leads to inaccurate and semantically inconsistent masks. In particular, without the contrastive loss, predictions become spatially fragmented and less coherent.

We further ablate the proposed multi-scale strategy by training with only a single scale level (``only coarse scale" or ``only fine scale"). As reported in Tab.~\ref{tab:ablation}, using a single scale consistently underperforms the full model, indicating that multi-scale supervision is crucial for handling objects with diverse sizes and granularities. Qualitative results in Fig.~\ref{fig:ablation_multi_scale} show that the coarse-only setting tends to over-group regions and inaccurate boundaries, whereas the fine-only setting often produces noisier and less stable masks. In contrast, combining multiple scales yields more coherent object-level segmentation while preserving fine details.

We also ablate the memory attention mechanism by removing the memory bank that accumulates masked CLIP features. Without the memory bank (``w/o memory"), mIoU drops by 21\%, and both FWIoU and accuracy decrease by 10\%, confirming the pivotal role of memory-based attention in semantic learning. This suggests that aggregating CLIP embeddings across frames provides more consistent semantic guidance than relying only on the current frame. Overall, these results verify our design choices and show that each component contributes significantly to the final performance.

\vspace{-3mm}

\begin{table}[!ht]
\centering
\resizebox{0.9\linewidth}{!}{
\begin{tabular}{l|cccc}
\toprule 
\textbf{Setting} & \textbf{mIoU}$\uparrow$ & \textbf{FWIoU}$\uparrow$ & \textbf{Acc}$\uparrow$ & \textbf{PSNR}$\uparrow$ \\
\midrule 
w/o contrastive loss & 0.477 & 0.628 & 0.697 & 34.02 \\
w/o regression loss & 0.616 & 0.660 & 0.705 & 34.30 \\
only coarse scale & 0.623 & 0.704 & 0.746 & 33.84 \\
only fine scale & 0.647 & 0.696 & 0.743 & 33.54 \\
w/o memory & 0.520 & 0.615 & 0.656 & 34.38 \\
\midrule 
\textbf{Ours (SAM1.0)} & \textbf{0.660} & \textbf{0.708} & \textbf{0.754} & \textbf{34.47} \\
\bottomrule 
\end{tabular}
}
\caption{Ablations. Our method achieves better semantic segmentation and reconstruction performance, demonstrating the importance of each component in our framework.}
\label{tab:ablation}
\vspace{-5mm}
\end{table}

\section{CONCLUSION}
We introduced \textit{OpenMonoGS-SLAM}, a novel semantic SLAM framework capable of open-set segmentation under a monocular setup. By integrating visual foundation models (VFMs) such as MASt3R, SAM, and CLIP, our system achieves robust generalization across diverse tasks, enabling accurate monocular camera tracking and mapping, while providing rich semantic understanding in open-world environments. Furthermore, our method operates without any depth input or 3D semantic ground truth, relying entirely on self-supervised learning objectives to guide both geometric reconstruction and semantic understanding. Experimental results demonstrate that OpenMonoGS-SLAM achieves superior performance compared to existing baselines, highlighting the promise of integrating VFMs into SLAM. However, a current limitation lies in the lack of robustness to dynamic scenes, inherited from the MASt3R backbone. This opens future directions for leveraging models like~\cite{zhang2024monst3r}, which are trained on dynamic environments, to enable even more general and wild-scene applicability.




\bibliographystyle{IEEEtran}          
\bibliography{IEEEabrv,refs}          

@inproceedings{keetha2024splatam,
  title={Splatam: Splat track \& map 3d gaussians for dense rgb-d slam},
  author={Keetha, Nikhil and Karhade, Jay and Jatavallabhula, Krishna Murthy and Yang, Gengshan and Scherer, Sebastian and Ramanan, Deva and Luiten, Jonathon},
  booktitle={Proceedings of the IEEE/CVF Conference on Computer Vision and Pattern Recognition},
  pages={21357--21366},
  year={2024}
}

@inproceedings{ha2024rgbd,
  title={Rgbd gs-icp slam},
  author={Ha, Seongbo and Yeon, Jiung and Yu, Hyeonwoo},
  booktitle={European Conference on Computer Vision},
  pages={180--197},
  year={2024},
  organization={Springer}
}

@inproceedings{yan2024gs,
  title={Gs-slam: Dense visual slam with 3d gaussian splatting},
  author={Yan, Chi and Qu, Delin and Xu, Dan and Zhao, Bin and Wang, Zhigang and Wang, Dong and Li, Xuelong},
  booktitle={Proceedings of the IEEE/CVF Conference on Computer Vision and Pattern Recognition},
  pages={19595--19604},
  year={2024}
}

@inproceedings{matsuki2024gaussian,
  title={Gaussian splatting slam},
  author={Matsuki, Hidenobu and Murai, Riku and Kelly, Paul HJ and Davison, Andrew J},
  booktitle={Proceedings of the IEEE/CVF Conference on Computer Vision and Pattern Recognition},
  pages={18039--18048},
  year={2024}
}

@inproceedings{hhuang2024photoslam,
	title = {Photo-SLAM: Real-time Simultaneous Localization and Photorealistic Mapping for Monocular, Stereo, and RGB-D Cameras},
	author = {Huang, Huajian and Li, Longwei and Cheng Hui and Yeung, Sai-Kit},
	booktitle = {Proceedings of the IEEE/CVF Conference on Computer Vision and Pattern Recognition},
	year = {2024}
}

@article{wen2025scaffold,
  title={Scaffold-slam: Structured 3d gaussians for simultaneous localization and photorealistic mapping},
  author={Wen, Tianci and Liu, Zhiang and Lu, Biao and Fang, Yongchun},
  journal={arXiv preprint arXiv:2501.05242},
  year={2025}
}

@inproceedings{li2024sgs,
  title={Sgs-slam: Semantic gaussian splatting for neural dense slam},
  author={Li, Mingrui and Liu, Shuhong and Zhou, Heng and Zhu, Guohao and Cheng, Na and Deng, Tianchen and Wang, Hongyu},
  booktitle={European Conference on Computer Vision},
  pages={163--179},
  year={2024},
  organization={Springer}
}

@article{li2024hier,
  title={Hier-SLAM: Scaling-up Semantics in SLAM with a Hierarchically Categorical Gaussian Splatting},
  author={Li, Boying and Cai, Zhixi and Li, Yuan-Fang and Reid, Ian and Rezatofighi, Hamid},
  journal={arXiv preprint arXiv:2409.12518},
  year={2024}
}

@inproceedings{kirillov2023segment,
  title={Segment anything},
  author={Kirillov, Alexander and Mintun, Eric and Ravi, Nikhila and Mao, Hanzi and Rolland, Chloe and Gustafson, Laura and Xiao, Tete and Whitehead, Spencer and Berg, Alexander C and Lo, Wan-Yen and others},
  booktitle={Proceedings of the IEEE/CVF international conference on computer vision},
  pages={4015--4026},
  year={2023}
}

@article{ren2024grounded,
  title={Grounded sam: Assembling open-world models for diverse visual tasks},
  author={Ren, Tianhe and Liu, Shilong and Zeng, Ailing and Lin, Jing and Li, Kunchang and Cao, He and Chen, Jiayu and Huang, Xinyu and Chen, Yukang and Yan, Feng and others},
  journal={arXiv preprint arXiv:2401.14159},
  year={2024}
}

@inproceedings{radford2021learning,
  title={Learning transferable visual models from natural language supervision},
  author={Radford, Alec and Kim, Jong Wook and Hallacy, Chris and Ramesh, Aditya and Goh, Gabriel and Agarwal, Sandhini and Sastry, Girish and Askell, Amanda and Mishkin, Pamela and Clark, Jack and others},
  booktitle={International conference on machine learning},
  pages={8748--8763},
  year={2021},
  organization={PmLR}
}

@inproceedings{leroy2024grounding,
  title={Grounding image matching in 3d with mast3r},
  author={Leroy, Vincent and Cabon, Yohann and Revaud, J{\'e}r{\^o}me},
  booktitle={European Conference on Computer Vision},
  pages={71--91},
  year={2024},
  organization={Springer}
}

@inproceedings{wang2025vggt,
  title={Vggt: Visual geometry grounded transformer},
  author={Wang, Jianyuan and Chen, Minghao and Karaev, Nikita and Vedaldi, Andrea and Rupprecht, Christian and Novotny, David},
  booktitle={Proceedings of the Computer Vision and Pattern Recognition Conference},
  pages={5294--5306},
  year={2025}
}

@article{yang2025opengs,
  title={OpenGS-SLAM: Open-Set Dense Semantic SLAM with 3D Gaussian Splatting for Object-Level Scene Understanding},
  author={Yang, Dianyi and Gao, Yu and Wang, Xihan and Yue, Yufeng and Yang, Yi and Fu, Mengyin},
  journal={arXiv preprint arXiv:2503.01646},
  year={2025}
}

@inproceedings{kerr2023lerf,
  title={Lerf: Language embedded radiance fields},
  author={Kerr, Justin and Kim, Chung Min and Goldberg, Ken and Kanazawa, Angjoo and Tancik, Matthew},
  booktitle={Proceedings of the IEEE/CVF international conference on computer vision},
  pages={19729--19739},
  year={2023}
}

@inproceedings{cen2025segment,
  title={Segment any 3d gaussians},
  author={Cen, Jiazhong and Fang, Jiemin and Yang, Chen and Xie, Lingxi and Zhang, Xiaopeng and Shen, Wei and Tian, Qi},
  booktitle={Proceedings of the AAAI Conference on Artificial Intelligence},
  volume={39},
  pages={1971--1979},
  year={2025}
}

@inproceedings{zou20253d,
  title={3D-SPATIAL MULTIMODAL MEMORY},
  author={Zou, Xueyan and Song, Yuchen and Qiu, Ri-Zhao and Peng, Xuanbin and Ye, Jianglong and Liu, Sifei and Wang, Xiaolong},
  booktitle={The Thirteenth International Conference on Learning Representations},
  year={2025}
}

@article{ye2025gsplat,
  title={gsplat: An open-source library for Gaussian splatting},
  author={Ye, Vickie and Li, Ruilong and Kerr, Justin and Turkulainen, Matias and Yi, Brent and Pan, Zhuoyang and Seiskari, Otto and Ye, Jianbo and Hu, Jeffrey and Tancik, Matthew and others},
  journal={Journal of Machine Learning Research},
  volume={26},
  number={34},
  pages={1--17},
  year={2025}
}

@inproceedings{zhi2021place,
  title={In-place scene labelling and understanding with implicit scene representation},
  author={Zhi, Shuaifeng and Laidlow, Tristan and Leutenegger, Stefan and Davison, Andrew J},
  booktitle={Proceedings of the IEEE/CVF International Conference on Computer Vision},
  pages={15838--15847},
  year={2021}
}

@article{straub2019replica,
  title={The replica dataset: A digital replica of indoor spaces},
  author={Straub, Julian and Whelan, Thomas and Ma, Lingni and Chen, Yufan and Wijmans, Erik and Green, Simon and Engel, Jakob J and Mur-Artal, Raul and Ren, Carl and Verma, Shobhit and others},
  journal={arXiv preprint arXiv:1906.05797},
  year={2019}
}

@inproceedings{li2024dns,
  title={Dns-slam: Dense neural semantic-informed slam},
  author={Li, Kunyi and Niemeyer, Michael and Navab, Nassir and Tombari, Federico},
  booktitle={2024 IEEE/RSJ International Conference on Intelligent Robots and Systems (IROS)},
  pages={7839--7846},
  year={2024},
  organization={IEEE}
}

@inproceedings{caron2021dino,
  title={Emerging properties in self-supervised vision transformers},
  author={Caron, Mathilde and Touvron, Hugo and Misra, Ishan and J{\'e}gou, Herv{\'e} and Mairal, Julien and Bojanowski, Piotr and Joulin, Armand},
  booktitle={Proceedings of the IEEE/CVF international conference on computer vision},
  pages={9650--9660},
  year={2021}
}

@article{weinzaepfel2022croco,
  title={Croco: Self-supervised pre-training for 3d vision tasks by cross-view completion},
  author={Weinzaepfel, Philippe and Leroy, Vincent and Lucas, Thomas and Br{\'e}gier, Romain and Cabon, Yohann and Arora, Vaibhav and Antsfeld, Leonid and Chidlovskii, Boris and Csurka, Gabriela and Revaud, J{\'e}r{\^o}me},
  journal={Advances in Neural Information Processing Systems},
  volume={35},
  pages={3502--3516},
  year={2022}
}

@inproceedings{jiang2024omniglue,
  title={Omniglue: Generalizable feature matching with foundation model guidance},
  author={Jiang, Hanwen and Karpur, Arjun and Cao, Bingyi and Huang, Qixing and Araujo, Andr{\'e}},
  booktitle={Proceedings of the IEEE/CVF Conference on Computer Vision and Pattern Recognition},
  pages={19865--19875},
  year={2024}
}

@inproceedings{edstedt2024roma,
  title={Roma: Robust dense feature matching},
  author={Edstedt, Johan and Sun, Qiyu and B{\"o}kman, Georg and Wadenb{\"a}ck, M{\aa}rten and Felsberg, Michael},
  booktitle={Proceedings of the IEEE/CVF Conference on Computer Vision and Pattern Recognition},
  pages={19790--19800},
  year={2024}
}

@Article{kerbl3Dgaussians,
      author       = {Kerbl, Bernhard and Kopanas, Georgios and Leimk{\"u}hler, Thomas and Drettakis, George},
      title        = {3D Gaussian Splatting for Real-Time Radiance Field Rendering},
      journal      = {ACM Transactions on Graphics},
      number       = {4},
      volume       = {42},
      month        = {July},
      year         = {2023},
      url          = {https://repo-sam.inria.fr/fungraph/3d-gaussian-splatting/}
}

@inproceedings{zhou2024feature,
  title={Feature 3dgs: Supercharging 3d gaussian splatting to enable distilled feature fields},
  author={Zhou, Shijie and Chang, Haoran and Jiang, Sicheng and Fan, Zhiwen and Zhu, Zehao and Xu, Dejia and Chari, Pradyumna and You, Suya and Wang, Zhangyang and Kadambi, Achuta},
  booktitle={Proceedings of the IEEE/CVF Conference on Computer Vision and Pattern Recognition},
  pages={21676--21685},
  year={2024}
}

@inproceedings{ye2024gaussian,
  title={Gaussian grouping: Segment and edit anything in 3d scenes},
  author={Ye, Mingqiao and Danelljan, Martin and Yu, Fisher and Ke, Lei},
  booktitle={European conference on computer vision},
  pages={162--179},
  year={2024},
  organization={Springer}
}

@inproceedings{schoenberger2016sfm,
    author={Sch\"{o}nberger, Johannes Lutz and Frahm, Jan-Michael},
    title={Structure-from-Motion Revisited},
    booktitle={Conference on Computer Vision and Pattern Recognition (CVPR)},
    year={2016},
}

@inproceedings{schoenberger2016mvs,
    author={Sch\"{o}nberger, Johannes Lutz and Zheng, Enliang and Pollefeys, Marc and Frahm, Jan-Michael},
    title={{Pixelwise View Selection for Unstructured Multi-View Stereo}},
    booktitle={European Conference on Computer Vision (ECCV)},
    year={2016}
}

@article{mildenhall2021nerf,
  title={Nerf: Representing scenes as neural radiance fields for view synthesis},
  author={Mildenhall, Ben and Srinivasan, Pratul P and Tancik, Matthew and Barron, Jonathan T and Ramamoorthi, Ravi and Ng, Ren},
  journal={Communications of the ACM},
  volume={65},
  number={1},
  pages={99--106},
  year={2021},
  publisher={ACM New York, NY, USA}
}

@inproceedings{murai2024_mast3rslam,
    title={{MASt3R-SLAM}: Real-Time Dense {SLAM} with {3D} Reconstruction Priors},
    author={Murai, Riku and Dexheimer, Eric and Davison, Andrew J.},
    booktitle={Conference on Computer Vision and Pattern Recognition (CVPR)},
    year={2025},
}

@article{davison2018futuremapping,
  title={FutureMapping: The computational structure of spatial AI systems},
  author={Davison, Andrew J},
  journal={arXiv preprint arXiv:1803.11288},
  year={2018}
}

@inproceedings{zhang2018lpips,
  title={The unreasonable effectiveness of deep features as a perceptual metric},
  author={Zhang, Richard and Isola, Phillip and Efros, Alexei A and Shechtman, Eli and Wang, Oliver},
  booktitle={Proceedings of the IEEE conference on computer vision and pattern recognition},
  pages={586--595},
  year={2018}
}

@article{wang2004ssim,
  title={Image quality assessment: from error visibility to structural similarity},
  author={Wang, Zhou and Bovik, Alan C and Sheikh, Hamid R and Simoncelli, Eero P},
  journal={IEEE transactions on image processing},
  volume={13},
  number={4},
  pages={600--612},
  year={2004},
  publisher={IEEE}
}

@article{zhang2024monst3r,
  author    = {Zhang, Junyi and Herrmann, Charles and Hur, Junhwa and Jampani, Varun and Darrell, Trevor and Cole, Forrester and Sun, Deqing and Yang, Ming-Hsuan},
  title     = {MonST3R: A Simple Approach for Estimating Geometry in the Presence of Motion},
  journal   = {arXiv preprint arxiv:2410.03825},
  year      = {2024}
}

@INPROCEEDINGS{tum,
  author={Sturm, Jürgen and Engelhard, Nikolas and Endres, Felix and Burgard, Wolfram and Cremers, Daniel},
  booktitle={2012 IEEE/RSJ International Conference on Intelligent Robots and Systems}, 
  title={A benchmark for the evaluation of RGB-D SLAM systems}, 
  year={2012},
  volume={},
  number={},
  pages={573-580},
  doi={10.1109/IROS.2012.6385773}}

@inproceedings{dai2017scannet,
  title={Scannet: Richly-annotated 3d reconstructions of indoor scenes},
  author={Dai, Angela and Chang, Angel X and Savva, Manolis and Halber, Maciej and Funkhouser, Thomas and Nie{\ss}ner, Matthias},
  booktitle={Proceedings of the IEEE conference on computer vision and pattern recognition},
  pages={5828--5839},
  year={2017}
}

@article{davison2007monoslam,
  title={MonoSLAM: Real-time single camera SLAM},
  author={Davison, Andrew J and Reid, Ian D and Molton, Nicholas D and Stasse, Olivier},
  journal={IEEE transactions on pattern analysis and machine intelligence},
  volume={29},
  number={6},
  pages={1052--1067},
  year={2007},
  publisher={IEEE}
}

@article{mur2015orb,
  title={ORB-SLAM: A versatile and accurate monocular SLAM system},
  author={Mur-Artal, Raul and Montiel, Jose Maria Martinez and Tardos, Juan D},
  journal={IEEE transactions on robotics},
  volume={31},
  number={5},
  pages={1147--1163},
  year={2015},
  publisher={IEEE}
}

@inproceedings{engel2014lsd,
  title={LSD-SLAM: Large-scale direct monocular SLAM},
  author={Engel, Jakob and Sch{\"o}ps, Thomas and Cremers, Daniel},
  booktitle={European conference on computer vision},
  pages={834--849},
  year={2014},
  organization={Springer}
}

@inproceedings{sandstrom2025splat,
  title={Splat-slam: Globally optimized rgb-only slam with 3d gaussians},
  author={Sandstr{\"o}m, Erik and Zhang, Ganlin and Tateno, Keisuke and Oechsle, Michael and Niemeyer, Michael and Zhang, Youmin and Patel, Manthan and Van Gool, Luc and Oswald, Martin and Tombari, Federico},
  booktitle={Proceedings of the Computer Vision and Pattern Recognition Conference},
  pages={1680--1691},
  year={2025}
}

\end{document}